\documentclass{article}

\usepackage{arxiv}

\usepackage[utf8]{inputenc} 
\usepackage[T1]{fontenc}    
\usepackage{hyperref}       
\usepackage{url}            
\usepackage{booktabs}       
\usepackage{amsmath}
\usepackage{amsfonts}       
\usepackage{nicefrac}       
\usepackage{microtype}      
\usepackage{graphicx}
\usepackage{subcaption}
\graphicspath{{./images/}{./materiels/Chap4/}}

\title{Toward an Operational GNN-Based Multimesh Surrogate for Fast Flood Forecasting}

\author{
  Valentin Mercier \\
  Université de Toulouse, INP, IRIT, Toulouse, France\\
  \texttt{valentin.mercier@toulouse-inp.fr.}  \\
  \And
  Serge Gratton \\
  Université de Toulouse, INP, IRIT, Toulouse, France\\
  \texttt{serge.gratton@toulouse-inp.fr.} \\
  \And 
  Corentin Lapeyre \\
  Researcher Engagement, Nvidia\\
  \texttt{clapeyre@nvidia.com.} \\
  \And 
  Gwenaël CHEVALLET \\
  Project Manager, BRL Ingénierie\\
  \texttt{Gwenael.Chevallet@brl.fr.} \\
}
\begin{document}
\maketitle

\begin{abstract}
Operational flood forecasting still relies on high-fidelity two-dimensional hydraulic solvers, but their runtime can be prohibitive for rapid decision support on large urban floodplains. In parallel, AI-based surrogate models have shown strong potential in several areas of computational physics for accelerating otherwise expensive high-fidelity simulations. We address this issue on the lower T\^et River (France), starting from a production-grade Telemac2D model defined on a high-resolution unstructured finite-element mesh with more than $4\times 10^5$ nodes. From this setup, we build a learning-ready database of synthetic but operationally grounded flood events covering several representative hydrograph families and peak discharges.

On top of this database, we develop a graph-neural surrogate based on projected meshes and multimesh connectivity. The projected-mesh strategy keeps training tractable while preserving high-fidelity supervision from the original Telemac simulations, and the multimesh construction enlarges the effective spatial receptive field without increasing network depth. We further study the effect of an explicit discharge feature $Q(t)$ and of pushforward training for long autoregressive rollouts.

The experiments show that conditioning on $Q(t)$ is essential in this boundary-driven setting, that multimesh connectivity brings additional gains once the model is properly conditioned, and that pushforward further improves rollout stability. Among the tested configurations, the combination of $Q(t)$, multimesh connectivity, and pushforward provides the best overall results. These gains are observed both on hydraulic variables over the surrogate mesh and on inundation maps interpolated onto a common $25\,\mathrm{m}$ regular grid and compared against the original high-resolution Telemac solution. On the studied case, the learned surrogate produces 6-hour predictions in about $0.4\,\mathrm{s}$ on a single NVIDIA A100 GPU, compared with about $180\,\mathrm{min}$ on 56 CPU cores for the reference simulation. These results support graph-based surrogates as practical complements to industrial hydraulic solvers for operational flood mapping.
\end{abstract}

\section{Introduction}
Floods are among the most destructive natural hazards, affecting billions of people worldwide and causing major economic losses \cite{STATFLOOD}. In operational river basins, the key product is often a flood hazard map: a spatially resolved estimate of water depth (and sometimes velocity) over the floodplain, updated fast enough to support emergency response. In practice, this product is consumed as a cartographic layer on a common spatial support rather than as raw values attached to solver nodes.

\paragraph*{Hydrodynamic modelling}
Two-dimensional hydrodynamic solvers based on the Saint-Venant (shallow-water) equations remain the reference approach for river flood prediction. They require consistent initial and boundary conditions, and a spatial discretisation fine enough to represent both the river channel and floodplain flow paths. In practice, this leads to highly inhomogeneous unstructured meshes refined along levees, roads, and urban obstacles; the resulting spread of edge lengths on the T\^et case is illustrated in Figure~\ref{fig:edge_length_dist}. Industrial and engineering workflows therefore still rely on mature hydraulic solvers such as HEC-RAS \cite{brunner1997hec} and Telemac2D \cite{hervouet}. In particular, Telemac2D combines a method of characteristics for advection with a finite-element formulation for propagation, diffusion, and source terms, and it offers a flexible set of boundary conditions (rigid walls and liquid boundaries).
For river applications, the upstream forcing is commonly provided as a discharge hydrograph, while the downstream boundary is given as a stage (sea level or rating curve). 

\paragraph*{Why machine-learning surrogates}

In this article, we consider the operational case of flood forecasting in the urbanised area of the River Têt in France. As detailed in Section~\ref{sec:data_construction}, accurately predicting floods in this area of about 100{,}000 inhabitants requires an unstructured mesh with roughly $4\times10^5$ nodes. Simulating a full 40-hour event with this setup can take up to 10 hours on 56 CPU cores. While this computational cost is acceptable for offline engineering studies, it can be prohibitive for time-critical forecasting. Machine-learning surrogates can complement the numerical solver by shifting most of the cost offline: once trained, they enable fast approximate predictions that can be converted into decision-oriented inundation maps in near real time.

\paragraph*{Existing neural approaches for floods.}
Deep-learning approaches for flood mapping have been reviewed extensively by Bentivoglio et al. \cite{hess-26-4345-2022}. Many early surrogates assume a structured raster representation and a fixed domain. For instance, Chu et al. \cite{Chu2020} propose an MLP-based emulation framework trained on 2D simulations on a 20\,m grid and able to predict water depths several hours ahead. Kao et al. \cite{kao2021fusing} combine a stacked autoencoder with an LSTM to forecast water heights on a 40\,m grid, using past events and additional event descriptors. While promising, such approaches can require large amounts of training data, they often do not transfer easily across geometries, and they can become difficult to scale when many local models are needed.
Recent sequence models extend this idea to predicting full flood maps: FloodSformer \cite{PIANFORINI2024131169} couples a convolutional encoder--decoder with a transformer-based video predictor to generate long sequences of dam-break water-depth maps on regular grids. Its runtime is compatible with emergency response, but the setting still relies on structured inputs and simplified boundary conditions, and autoregressive rollouts remain sensitive to error accumulation.

Physics-informed neural solvers provide another direction. FloodCast/GeoPINS \cite{XU2024122162} combine geometry adaptation with Fourier Neural Operators and enforce mass and momentum conservation through a hybrid data/physics formulation. These models can handle large domains and include boundary conditions, but they also illustrate common limitations in flood applications: long-horizon accumulation errors, smoothing of fast dynamics, and reduced flexibility when boundary conditions must be hard-coded.

\paragraph*{Graph neural networks and finite-volume informed designs.}
Graph neural networks are natural candidates when the underlying solver operates on a mesh. They represent the computational domain as a graph and learn local update rules through message passing, which has proven effective for CFD surrogates on unstructured meshes \cite{meshgraphnet}. For floods, Bentivoglio et al. introduce SWE--GNN \cite{Bentivoglio2023}, explicitly leveraging the analogy between graph message passing and finite-volume flux computations. Each cell is a node, edges connect neighboring cells, and a learned local function predicts flux-like exchanges driven by hydraulic gradients. With multi-step training and curriculum strategies, the model achieves accurate long-horizon predictions and large speed-ups over the numerical solver on dam-break datasets.
However, the original demonstrations remain based on small, regular meshes and simplified boundary conditions (e.g., constant breach inflow). Follow-up work proposes a multi-scale variant with ghost cells to handle time-varying boundaries and rotation-invariant features \cite{egusphere-2024-2621}, but current tests are still limited to domains with tens of thousands of cells.

Recent NVIDIA-affiliated work proposes HydroGraphNet \cite{taghizadeh2025hydrographnet}, a physics-informed GNN that enforces mass conservation during training (without expensive automatic differentiation) and improves both stability and critical success index on major flood events. The model also incorporates techniques to mitigate autoregressive drift and includes interpretable components. Its ablation study further highlights the value of explicitly conditioning the surrogate on global inflow information, which directly informs our use of the discharge feature $Q(t)$. These results confirm that combining mesh-based learning with conservation constraints is a promising route for operational flood forecasting.

\paragraph*{Where our work fits: finite elements, boundary conditions, and inundation priorities.}
Most physics-informed GNN flood surrogates above build on finite-volume discretizations, where state variables are cell-centered and conservation errors can be written directly in terms of cell volumes and face fluxes \cite{darwish2016finite}. Our operational database, in contrast, is produced by Telemac2D on unstructured finite-element meshes \cite{bathe2006finite}: the state is nodal, fluxes are implicit in a weak formulation, and liquid boundary conditions are imposed on node sets through combinations of prescribed/free depth and velocity \cite{hervouet}. In particular, the upstream forcing is a time-varying hydrograph, and the downstream condition represents the sea level. This boundary-driven setting changes how inputs must be encoded and makes finite-volume flux penalties not directly reusable.
Rather than forcing an approximate finite-volume conservation penalty on nodal FEM outputs, we exploit the high-fidelity Telemac database and focus on graph representation of the problem using projection and multimesh technique. Accordingly, our objective is not only to predict nodal hydraulic states on a tractable mesh representation, but also to preserve the quality of the final inundation maps obtained after interpolation onto a common fine grid.

\paragraph*{Contributions and outline.}
Building on an operational flood database from the industrial sector, this paper makes two key contributions. First, we introduce a graph-based surrogate strategy tailored to this boundary-driven finite-element setting, combining projected meshes with multimesh connectivity in order to enlarge the effective spatial receptive field while keeping learning tractable. Second, we evaluate the resulting models not only on hydraulic variables over the surrogate mesh, but also on inundation maps interpolated onto a common regular grid and compared against the original high-resolution Telemac solution. The remainder of the paper describes the construction of the database, the projected and multimesh graph representations, the training protocol and ablation experiments, and finally the quantitative and qualitative evaluation results.

\section{Task description and data construction}
\label{sec:data_construction}
We build the learning-ready dataset in four stages: mesh generation, specification of initial and boundary conditions, synthesis of representative hydrographs, and execution of the hydrodynamic simulations that populate the database. Each stage leverages domain expertise gathered during the BRLi operational studies on the Têt River, ensuring that the resulting data inherits the same level of fidelity as the production-grade model.
The Têt basin is also a relevant operational test case because flood-prone urban areas are located close to the riverbanks, so small errors in water level or flood extent can translate into large differences in exposed assets.

\subsection{Mesh generation}
\label{sec:mesh_generation}
To create a reliable dataset for the Têt River, we used a density map to build the computational mesh (Fig.~\ref{fig:density}).
The mesh was carefully designed to capture key features of the region:
\begin{itemize}
    \item \textbf{Main riverbed and embankments:} These zones have the highest density, with an average distance between nodes ranging from 5 to 10 metres.
    \item \textbf{Urban areas:} Zones near populated areas had a refined density to improve accuracy in regions most at risk of flooding.
    \item \textbf{General area:} Other parts of the study area have an average distance between nodes of 50 metres.
    \item \textbf{Critical overflow corridors:} Areas where the river can breach into secondary channels or reclaimed land were meshed at 5 metres to preserve the topology of protective levees and roads.
\end{itemize}
Softlines were used to refine the modeling of key structures, such as embankments near the riverbed, helping the model better capture flood behavior.
Using this configuration, we obtain a mesh of 412,844 nodes and 2,013,374 elements. This mesh is interpolated onto the topographical surveys conducted in the area to obtain the bed elevation and friction values.

\subsection{Initial and boundary conditions}
\label{sec:icbc}
To model the flood, upstream liquid boundary conditions are applied, downstream boundary conditions are used to model the sea, and rigid wall conditions are applied elsewhere.
Initially, our domain is empty. To avoid unrealistic transients, we first dug an artificial lake upstream and filled it with water.
We then run an initialization simulation with an upstream flow rate of $100 \, \mathrm{m}^3/\mathrm{s}$ and a downstream imposed level gradually rising to $1.5 \, \mathrm{m}$.
This simulation runs for $100 \, h$ until the domain stabilizes.

\subsection{Generating realistic floods}
\label{sec:hydrographs_generation}
To create a realistic database, we based our approach on a historical analysis of floods to generate hydrographs.
Specifically, we selected the 20 most significant historical floods in terms of peak discharge.
We then calculated the shape coefficient ($K_c$), which is determined between the peak of the hydrograph and the daily mean discharge.
This daily mean discharge is computed to be maximized over the 24-hour window containing the flood peak.
The higher the shape coefficient, the sharper the flood hydrograph, indicating a shorter duration and smaller volume.

\begin{figure}[t]
    \centering
    \begin{subfigure}{0.48\linewidth}
        \centering
        \includegraphics[width=\linewidth]{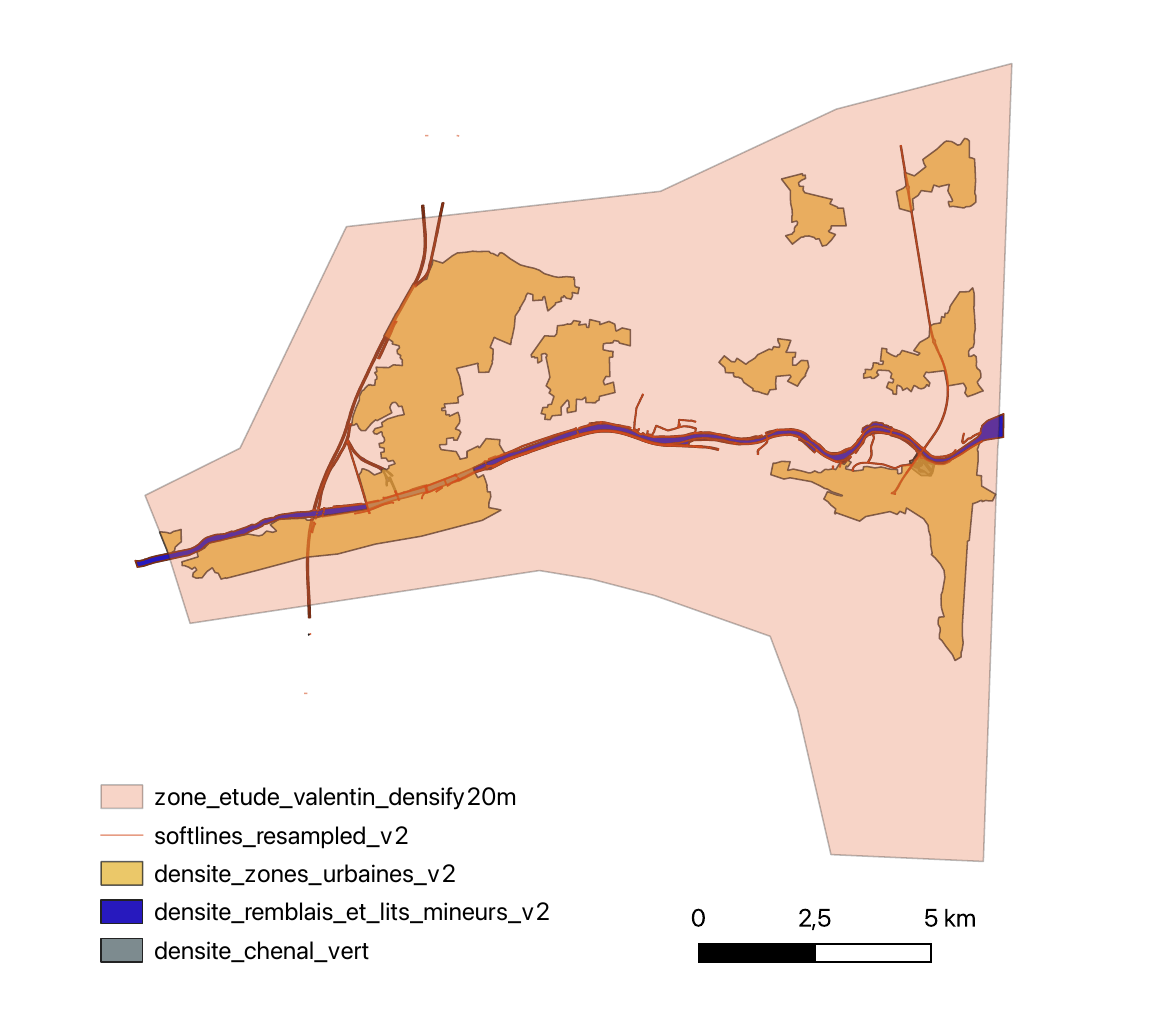}
        \caption{Mesh-density map guiding the unstructured discretisation of the Têt River floodplain.}
        \label{fig:density}
    \end{subfigure}
    \hfill
    \begin{subfigure}{0.48\linewidth}
        \centering
        \includegraphics[width=\linewidth]{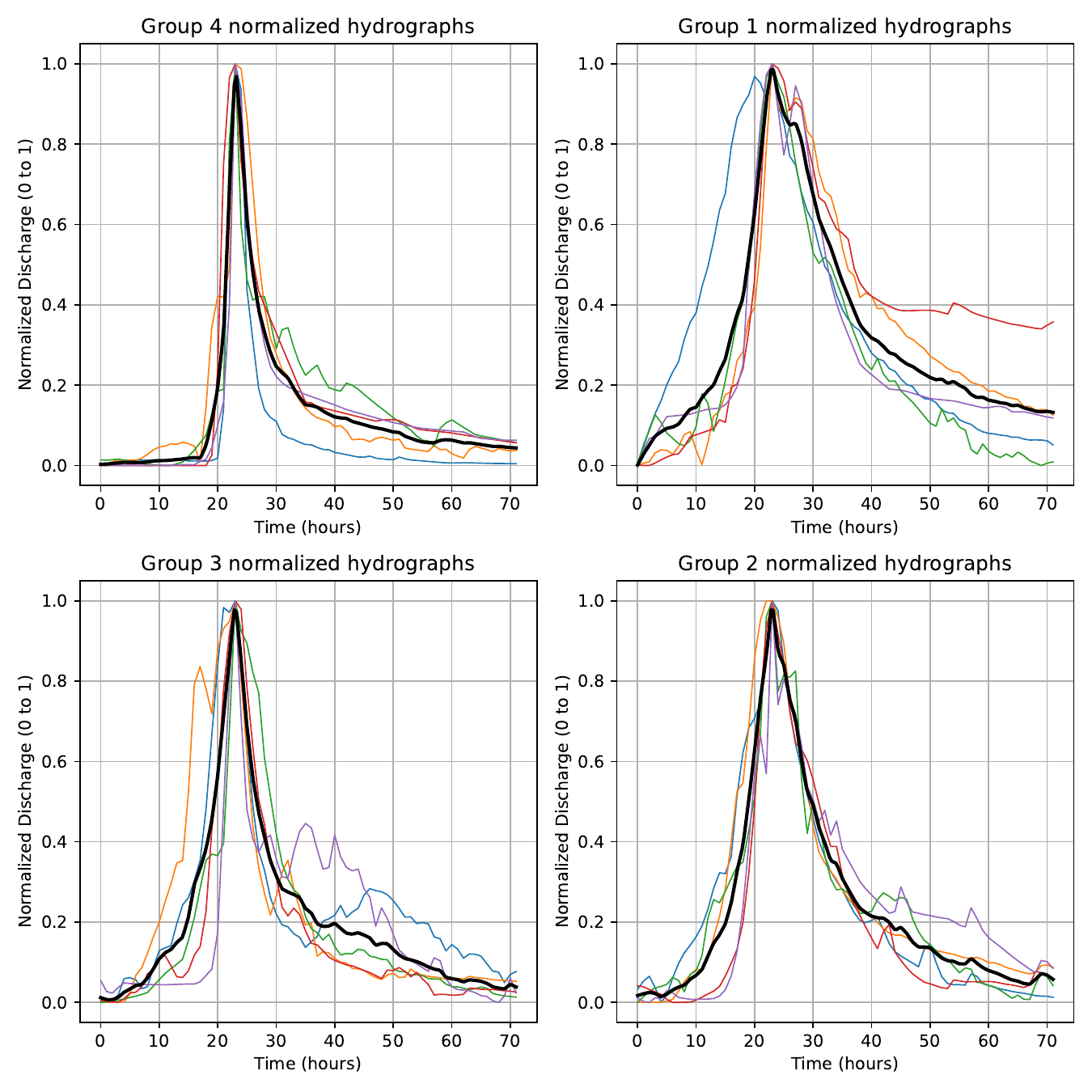}
        \caption{Representative hydrographs derived from the historical catalogue.}
        \label{fig:hydrographs}
    \end{subfigure}
    \caption{Overview of the dataset design: (a) spatial mesh-density strategy and (b) synthetic hydrograph families used to populate the database.}
    \label{fig:density_hydro_overview}
\end{figure}

We used this index to cluster the hydrograph types into four distinct groups.
Each group was then normalized and averaged, resulting in four representative hydrographs (Fig.~\ref{fig:hydrographs}).
Once these typical hydrographs were obtained, we scaled them by multiplying with peak discharges ranging between that of the 2020 flood ($1000 \, \mathrm{m}^3/\mathrm{s}$) and that of the 1940 flood ($3600 \, \mathrm{m}^3/\mathrm{s}$), starting the hydrographs at $100 \, \mathrm{m}^3/\mathrm{s}$ to match the initialization discharge.
This scaling was performed across 14 value intervals.

\subsection{Hydrodynamic simulations and stored variables}
\label{sec:simulations_storage}
Using the generated meshes and the defined initial and boundary conditions, we utilized Telemac2D.
As a result, we generated $14 \times 4 = 56$ hydrographs, each with a duration of 40 hours, forming our database.
We obtain a result in Selafin format \cite{hervouet}, which contains at each node the water height and velocities, as well as static bed elevation and friction values.
We chose to have an output every 30 minutes in our database.

\section{MeshGraphNet surrogate model}
\label{sec:meshgraphnet_model}
We model the flood dynamics on the unstructured mesh using the MeshGraphNet architecture \cite{meshgraphnet}. The goal is to learn an operator that maps the hydraulic state at time $t$ to the state at time $t+\Delta t$ (here $\Delta t = 30 \text{ min}$) on the choosen mesh.

\subsection{Graph representation}
\label{sec:graph_representation}
Each Telemac node is represented as a graph node. We split node features into:
\begin{itemize}
    \item \textbf{Dynamic variables:} water depth $h$ and horizontal velocities $(u,v)$ at the current time step.
    \item \textbf{Static variables:} bed elevation $z$ and friction parameter (Strickler coefficient), which do not change over time.
    \item \textbf{Boundary descriptors:} a one-hot encoding describing whether the node belongs to an inflow boundary, an imposed water elevation boundary, a wall boundary, or the interior.
\end{itemize}
Edges follow the mesh connectivity: two nodes are connected whenever they share a mesh edge. In practice, each mesh edge is stored twice, as two directed edges $(i,j)$ and $(j,i)$: the adjacency is symmetric, but messages are directional. For each directed edge $(i,j)$, we use geometric edge features based on the relative displacement in the horizontal plane, e.g. $(x_j-x_i,\, y_j-y_i)$ and its norm. The reverse edge therefore carries the opposite displacement with the same distance, which lets message passing distinguish update directions. This directional encoding is important on our highly inhomogeneous unstructured meshes, where edge lengths and node degrees vary strongly across the domain (by orders of magnitude between the refined riverbed and the floodplain).
We purposely rely on relative geometry rather than absolute coordinates: the network sees where a neighbor is \emph{with respect to} a node, which is the quantity required to learn local update rules on unstructured meshes.
In practice, dynamic variables, static variables, and edge features are normalized (mean/variance) using training-set statistics to ease optimisation.

\subsection{Encoder--processor--decoder architecture}
\label{sec:meshgraphnet_architecture}
MeshGraphNet uses two encoders to map node and edge features to a latent space, a processor made of $L$ message-passing blocks, and a decoder that outputs the predicted update of the dynamic variables. The encoders are multi-layer perceptrons applied independently to each node and each edge, and they lift heterogeneous inputs (hydraulic variables, terrain parameters, and geometric features) into a shared latent dimension.
In our setting, the node encoder receives the concatenation of dynamic variables (updated each step) and static/boundary descriptors (constant for a given mesh). The edge encoder receives the geometric edge features. All encoder, processor, and decoder parameters are shared across nodes, edges, and time steps, which enables training and inference on meshes with different sizes and resolutions.

The processor then alternates edge updates and node updates with residual connections. Denoting by $\mathbf{v}_i^\ell$ and $\mathbf{e}_{ij}^\ell$ the latent node and edge features at processor depth $\ell$, one typical message-passing block reads:
\begin{equation}
\mathbf{e}_{ij}^{\ell+1} = \mathbf{e}_{ij}^{\ell} + \phi_e\!\left([\mathbf{e}_{ij}^{\ell},\mathbf{v}_i^{\ell},\mathbf{v}_j^{\ell}]\right),\quad
\mathbf{v}_i^{\ell+1} = \mathbf{v}_i^{\ell} + \phi_v\!\left([\mathbf{v}_i^{\ell}, \sum_{j\in\mathcal{N}(i)} \mathbf{e}_{ij}^{\ell+1}]\right),
\end{equation}
where $\phi_e$ and $\phi_v$ are multi-layer perceptrons and $\mathcal{N}(i)$ denotes the neighbors of node $i$.
Repeating this block $L$ times increases the receptive field to $L$ hops on the graph.
The decoder maps the final latent node features to physical-space predictions. We predict increments $(\Delta h,\Delta u,\Delta v)$ rather than absolute values, and we advance the state by:
$
(h,u,v)^{t+\Delta t} = (h,u,v)^{t} + (\Delta h,\Delta u,\Delta v).
$

At inference time, the model is rolled out autoregressively: the predicted state at $t+\Delta t$ is fed back as input to predict $t+2\Delta t$, and so on. This makes the stability of the learned operator a central concern, especially in long hydrographs where small local errors can accumulate and alter inundation dynamics. During training, we therefore treat the model as a one-step update rule but evaluate it under multi-step rollouts, as this matches the operational use case (several hours of prediction from a given state and boundary forcing).

\subsection{Boundary conditions in the surrogate}
\label{sec:surrogate_boundary_conditions}
In practice, boundary nodes act as \emph{driver nodes}: their next-step values are injected into the graph inputs, and the model is not asked to learn updates on them. This separation avoids mixing boundary specification rules with the interior dynamics learned by message passing, while still allowing boundary information to propagate into the domain through the graph connectivity.
More precisely, nodes that belong to prescribed-discharge boundaries are provided with the imposed $(h,u,v)$ at the next step, while nodes that belong to imposed-stage boundaries are provided with the imposed next-step water depth. The model then predicts updates only for interior nodes (and for wall nodes), and the boundary updates are set to zero by construction. This is consistent with the operational workflow, where upstream and downstream forcings are known inputs during forecasting.

\subsection{Projection onto coarser meshes}
\label{sec:projection_coarse}
Flood surrogates based on message passing remain local: after $L$ processor layers, information has travelled only $L$ hops on the graph. On the production Telemac mesh, even $L\!\approx\!10$ corresponds to a very small physical radius compared with the kilometre-scale floodplain, which makes it difficult to capture long-range inundation patterns with a reasonable network depth. The simplest remedy is to operate on a coarser mesh, where each hop spans larger distances and can therefore propagate hydraulic information over a wider physical support, as illustrated by the shift toward longer edges in Figure~\ref{fig:edge_length_dist}.

For this reason, each Telemac simulation stored on the reference mesh is also projected onto a family of coarser meshes. Coarse meshes are generated with the same meshing software by relaxing the density map: we divide each density prescription by factors of $2$, $4$, $8$, $16$, and $32$ (equivalently, we multiply the target node spacing). This keeps the geometry consistent with the production setup while reducing the number of nodes to a scale that is tractable for learning.

Importantly, we do not run Telemac2D on these very coarse meshes, as this would degrade the hydrodynamic solution. Instead, we adopt the \emph{high-accuracy labels} principle of MultiScale MeshGraphNets \cite{fortunato2022multiscalemeshgraphnets}: we run the reference solver at the highest feasible resolution and interpolate its outputs onto the coarser meshes, so that coarse-resolution training targets inherit the fidelity of the fine simulation while exposing the network to a larger effective spatial context. We use the $\times 8$ projected dataset for the experiments reported in this paper, as it keeps graph sizes and training costs manageable. In this workflow, the projected mesh should be viewed as a tractable computational support for the surrogate, not as the final cartographic support used to assess inundation maps. A natural concern in this projection step is aliasing induced by spatial downsampling. In additional spectral checks on this dataset, we found little energy at the highest spatial frequencies, and the bilinear interpolation used to construct the projected targets also introduces some smoothing, so we did not apply any extra anti-aliasing filter.

\subsection{Multimesh graph construction}
\label{sec:multimesh_graph}
\begin{figure}[t]
    \centering
    \includegraphics[width=0.92\linewidth]{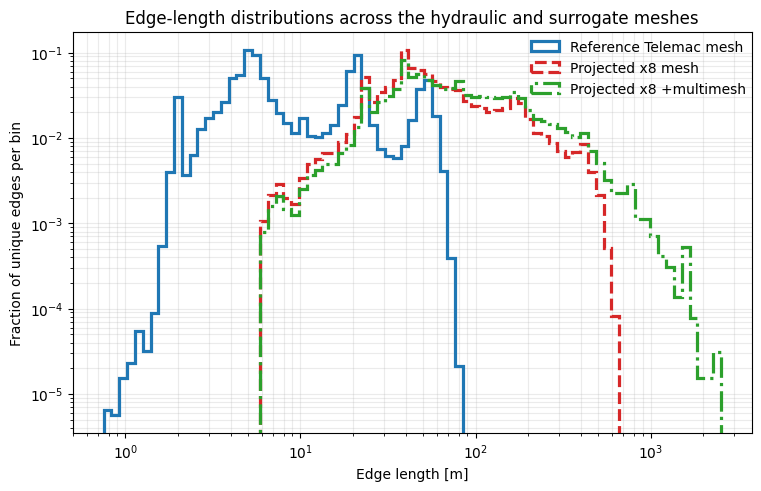}
    \caption{Distribution of unique edge lengths for the reference Telemac mesh, the projected $\times 8$ mesh, and the projected $\times 8$ + multimesh graph. The reference hydraulic mesh exhibits a broad distribution of edge lengths, reflecting its strongly inhomogeneous spatial resolution. Projection to the $\times 8$ mesh shifts the distribution toward larger local interaction scales, while multimesh connectivity adds a sparse tail of longer-range connections.}
    \label{fig:edge_length_dist}
\end{figure}

To further increase the effective receptive field without increasing the number of message-passing layers, we construct an augmented graph with additional long-range edges. For each mesh coarser than the $\times 8$ mesh, we match its nodes to the closest nodes in the original mesh (using a k-d tree) and add edges between the matched fine nodes. Aggregating the edges produced by several coarse levels yields a multi-scale connectivity pattern that allows information to travel across kilometres of the floodplain in a small number of message-passing steps while keeping the latent dimension and processor depth unchanged.
Operationally, this can be interpreted as transferring coarse mesh connectivity to the fine graph: each edge of a coarse auxiliary mesh becomes a shortcut between the corresponding nearest fine nodes. The union over several coarse levels provides a sparse set of long-range links that complements the native mesh adjacency (which is dominated by short edges in the refined riverbed); in Figure~\ref{fig:edge_length_dist}, this appears as the additional long-edge tail of the multimesh graph relative to the projected $\times 8$ mesh. 

\begin{table}[t]
    \centering
    \small
    \begin{tabular}{lrrrrrr}
        \toprule
        Graph & Nodes & Directed edges & Mean edge (m) & $r_{10}$ median (m) & $r_{10}$ mean (m) & $r_{10}$ max (m) \\
        \midrule
        Original mesh & 412{,}844 & 2{,}464{,}330 & 15.7 & 67 & 156 & 733 \\
        $\times 8$ mesh & 16{,}222 & 96{,}984 & 81.0 & 479 & 809 & 5{,}468 \\
        $\times 8$ + multimesh & 16{,}222 & 130{,}597 & 110.4 & 596 & 994 & 11{,}571 \\
        \bottomrule
    \end{tabular}
    \caption{Indicative 10-hop physical scale for different graph constructions. We report summary statistics across nodes of $r_{10}(i)=10\,\overline{\ell}_i$, where $\overline{\ell}_i$ is the mean outgoing edge length at node $i$ (all distances in metres).}
    \label{tab:hop_radius}
\end{table}

Figure~\ref{fig:edge_length_dist} complements Table~\ref{tab:hop_radius} by showing the full edge-length distributions behind these graph constructions. We report in Table~\ref{tab:hop_radius} the evolution of the distance covered in 10 hops (10 message-passing steps) across graph constructions using the indicative scale $r_{10}$ defined in the caption. On the original mesh, the median 10-hop scale is $r_{10}\approx 67$\,m (mean $156$\,m). On the $\times 8$ mesh, it increases to $r_{10}\approx 479$\,m (mean $809$\,m). Finally, adding multimesh shortcuts on top of the $\times 8$ mesh further expands the typical scale to $r_{10}\approx 596$\,m (mean $994$\,m) and increases the maximum value to about $11.6$\,km, illustrating the effect of sparse long-range edges on the physical receptive field.

\section{Training and evaluation protocol}
\label{sec:training_eval}
This paper focuses on developing an operationally motivated surrogate model for rapid flood forecasting on the River Têt (France), at decision-relevant lead times. The training and evaluation of the different model variants are therefore designed to reflect this setting. The split is performed at the flood-event level: 40 floods are used for training and 16 floods are kept held out for evaluation, with all rollout segments extracted from a given flood assigned to the same split. We first present the training procedure, which leads to a set of ablation experiments, and then the metrics used to quantify performance across multiple forecast horizons.
\subsection{Pushforward training}
\label{sec:pushforward}
Standard one-step training minimises an error on the predicted increment $(\Delta h,\Delta u,\Delta v)$ at time $t$.
However, at inference time the model is unrolled for multiple steps, and small biases may accumulate.
To reduce this train--test mismatch, we use a pushforward strategy, following autoregressive training ideas used in neural PDE solvers \cite{brandstetter2022messagepassingneuralpde}: during training, the next input state is stochastically chosen between the ground-truth state (teacher forcing) and the model-predicted state.
Denoting by $\hat{\mathbf{x}}^{t+1}$ the predicted next state (after boundary injection) and by $\mathbf{x}^{t+1}$ the ground truth, the rollout input is:
\begin{equation}
\tilde{\mathbf{x}}^{t+1} =
\begin{cases}
\mathbf{x}^{t+1}, & \text{with probability } p_{\mathrm{TF}}(e),\\
\hat{\mathbf{x}}^{t+1}, & \text{otherwise,}
\end{cases}
\end{equation}
where $e$ is the training epoch and $p_{\mathrm{TF}}$ decreases linearly from $1$ to $0$ over a warmup period (so the model gradually learns to recover from its own errors).
To keep training stable and memory-efficient, the reinjected state is detached, so we do not backpropagate through long rollouts.
\subsection{Adding the discharge as a global node feature}
\label{sec:global_discharge_feature}
Although the upstream discharge hydrograph $Q(t)$ is imposed at the boundary, it carries global information about the timing and magnitude of the event. Following the ablation study of HydroGraphNet \cite{taghizadeh2025hydrographnet}, which highlights the usefulness of explicitly conditioning flood surrogates on inflow information, we optionally augment the dynamic node input with $Q(t)$ at each time step. Concretely, we append the scalar discharge to the feature vector of every node (in addition to the boundary injection described above). This exogenous forcing signal conditions the surrogate on the current hydrological regime and can reduce the burden on message passing to propagate boundary information across long distances. We evaluate the impact of this feature in the ablation study.

\subsection{Main experiments}
\label{sec:main_experiments}
We consider six main training configurations, which we will refer to by their names in Table~\ref{table:experiments}. All experiments are run on the $\times 8$ projected dataset. We ablate (i) the addition of the global discharge feature $Q(t)$ (Section~\ref{sec:global_discharge_feature}), (ii) the use of multimesh connectivity (Section~\ref{sec:multimesh_graph}), and (iii) pushforward training (Section~\ref{sec:pushforward}), which we evaluate on top of the $Q(t)$ configuration. All runs are trained for 900 epochs; when enabled, pushforward is applied during the last 300 epochs, with a 150-epoch warmup. Because training is computationally expensive, this preprint does not include an extensive hyperparameter search or validation-based checkpoint selection: all reported results correspond to the final checkpoint.

\begin{table}[t]
    \centering
    \small
    \begin{tabular}{llcc}
        \toprule
        Name & Connectivity & $Q(t)$ & PF \\
        \midrule
        E1 & standard & no & no \\
        E2 & standard & yes & no \\
        E3 & standard & yes & yes \\
        E4 & multimesh & no & no \\
        E5 & multimesh & yes & no \\
        E6 & multimesh & yes & yes \\
        \bottomrule
    \end{tabular}
    \caption{Main experimental configurations. PF denotes pushforward training.}
    \label{table:experiments}
\end{table}

\subsection{Other training setups}
We train with the Adam optimiser using an initial learning rate of $5\times 10^{-4}$ and an exponential epoch-wise decay factor of $0.999995$. All training runs are performed using 8 NVIDIA A100 GPUs (80\,GB), distributed over 4 nodes.
\subsection{Metrics}
\label{sec:metrics}
We report two complementary families of metrics. First, we measure state-space errors on the projected mesh using denormalized errors on the hydraulic variables $(h,u,v)$. These projected targets are not produced by rerunning Telemac on the $\times 8$ mesh: they are obtained by interpolating the original high-resolution Telemac solution onto the projected support. Second, because the main operational product is an inundation map, we interpolate both the surrogate predictions and the reference Telemac solution from the original mesh onto a common fine regular grid and evaluate flood extent agreement on that grid.
For a threshold $h_\varepsilon$, the critical success index (CSI) is defined as
\begin{equation}
\mathrm{CSI} = \frac{\mathrm{TP}}{\mathrm{TP}+\mathrm{FP}+\mathrm{FN}},
\end{equation}
where TP, FP, and FN count flooded and non-flooded grid cells on the common fine grid. Unless otherwise stated, we use $h_\varepsilon = 5\,\mathrm{cm}$.

\section{Numerical Results}
All numerical results are reported on the 16 held-out flood events that are not seen during training. When we show time-resolved rollout (unrolling) curves, each value is the mean metric computed over these held-out floods at the corresponding lead time. Unless otherwise stated, inundation-map scores are computed after interpolating both predictions and reference solutions onto the same fine regular grid. We first assess the impact of the global discharge feature $Q(t)$, then the additional contribution of multimesh connectivity in the $Q(t)$-conditioned setting, then the effect of pushforward training, and finally the resulting fine-grid CSI at two operational depth thresholds.

\subsection{Ablation of the global discharge feature}
\label{sec:results_q}
\begin{figure}[t]
    \centering
    \includegraphics[width=\linewidth]{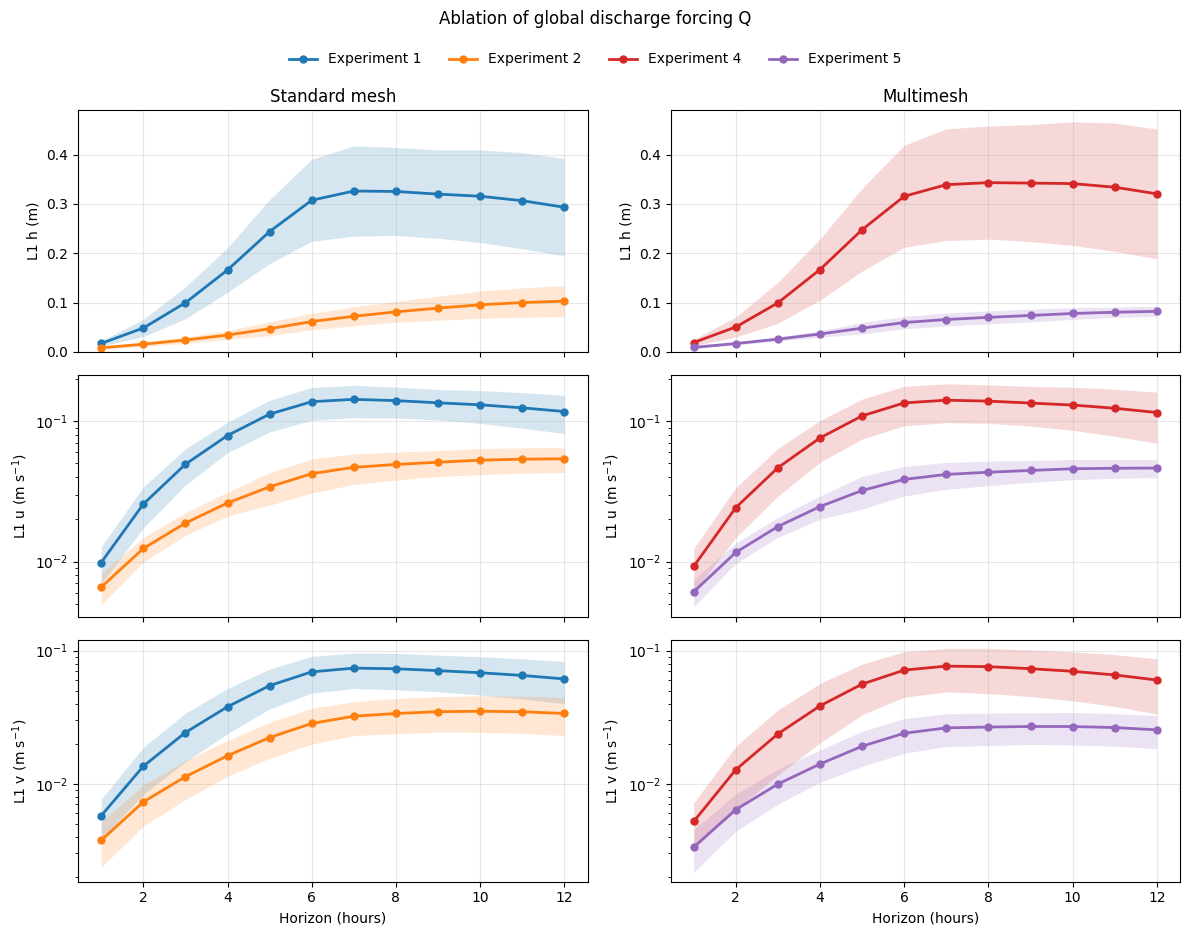}
    \caption{Ablation of the global discharge feature $Q(t)$ on the 16 held-out floods. Left: standard mesh (E1 vs E2). Right: multimesh (E4 vs E5). Solid lines show the mean held-out $L1$ error at each lead time and shaded areas indicate one standard deviation.}
    \label{fig:ablation_q}
\end{figure}

Figure~\ref{fig:ablation_q} compares the effect of adding the broadcast hydrograph feature $Q(t)$ when pushforward training is disabled. On the standard mesh, E2 consistently improves over E1 for water depth and both velocity components across the whole rollout horizon. The same trend is observed on the multimesh graph, where E5 markedly reduces the errors of E4. In our setting, multimesh connectivity alone is therefore not sufficient: without an explicit global forcing signal, the model struggles to recover the timing and magnitude of the event. Once $Q(t)$ is injected at every node, errors remain substantially lower, especially beyond 4--6 hours where autoregressive drift becomes more visible. This first ablation establishes $Q(t)$ as a necessary conditioning signal; the remaining comparisons therefore focus on $Q(t)$-augmented models.

\subsection{Ablation of multimesh connectivity}
\label{sec:results_multimesh}
\begin{figure}[t]
    \centering
    \includegraphics[width=\linewidth]{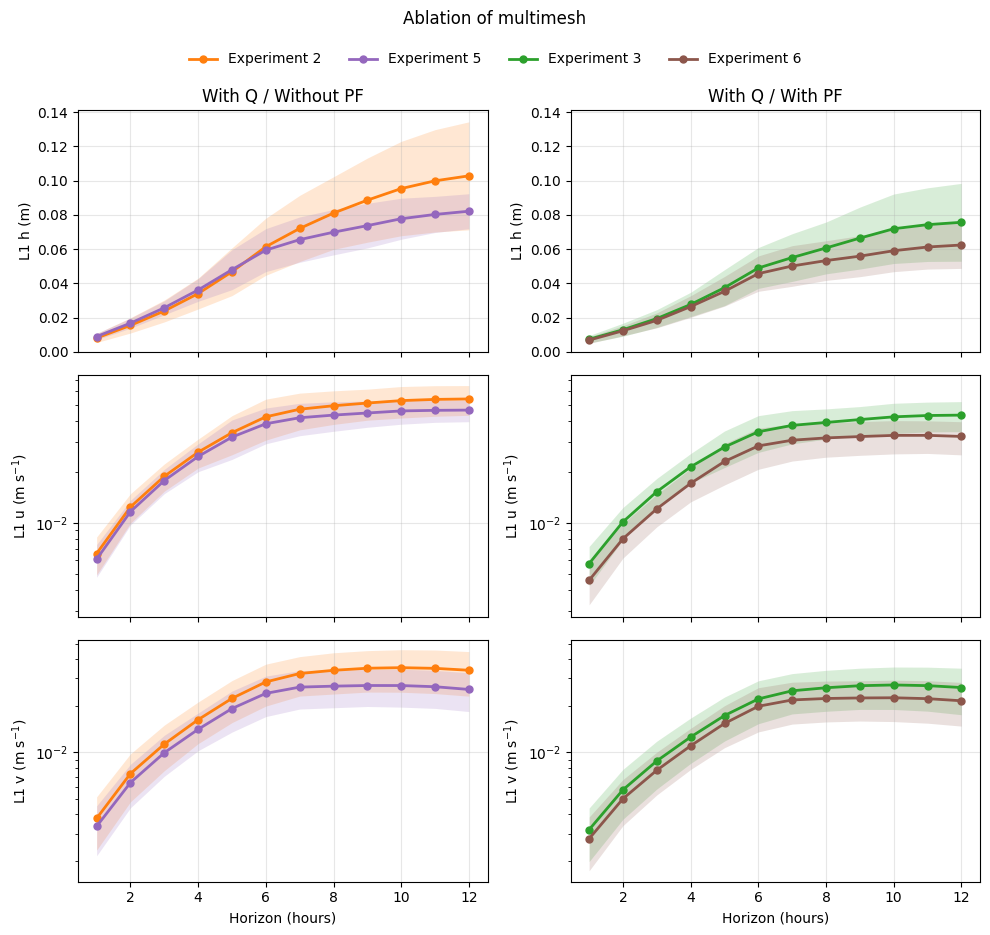}
    \caption{Ablation of multimesh connectivity on the 16 held-out floods, restricted to $Q(t)$-conditioned models. Left: no pushforward (E2 vs E5). Right: with pushforward (E3 vs E6). Solid lines show the mean held-out $L1$ error at each lead time and shaded areas indicate one standard deviation.}
    \label{fig:ablation_m}
\end{figure}

Figure~\ref{fig:ablation_m} isolates the effect of multimesh connectivity once the surrogate is conditioned on $Q(t)$. Without pushforward, E5 improves over E2, with the clearest gains on water depth and smaller but still consistent reductions on the velocity errors. When pushforward is also enabled, E6 improves over E3 on all three variables and at essentially all horizons. This indicates that the additional long-range edges bring useful spatial context beyond the discharge forcing alone, and that this benefit is more pronounced once rollout stability has been improved by pushforward training. Compared with the large jump observed when adding $Q(t)$, the multimesh gain is more moderate, but it remains systematic in the configurations that matter most for forecasting.

\subsection{Ablation of pushforward training}
\label{sec:results_pf}
\begin{figure}[t]
    \centering
    \includegraphics[width=\linewidth]{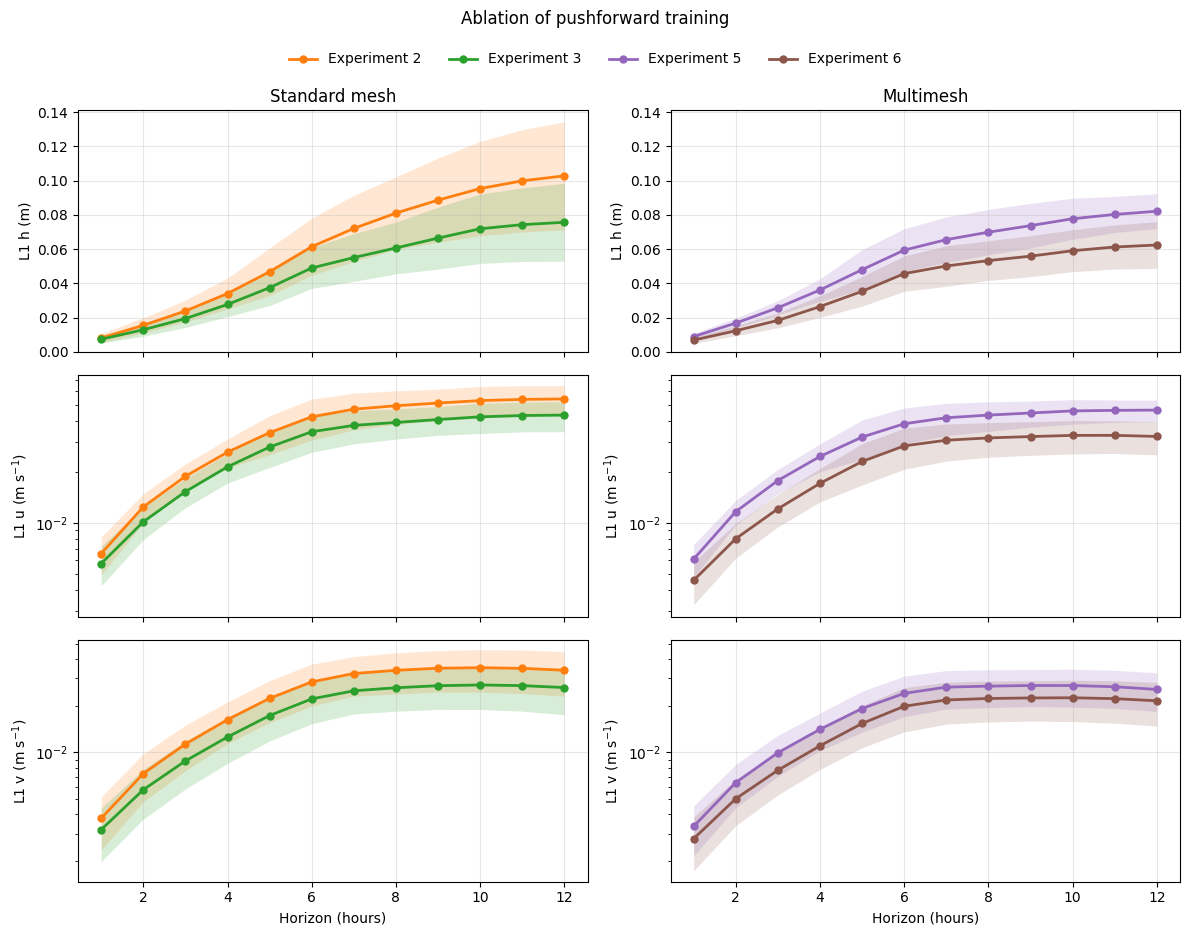}
    \caption{Ablation of pushforward training on the 16 held-out floods. Left: standard mesh with $Q(t)$ (E2 vs E3). Right: multimesh with $Q(t)$ (E5 vs E6). Solid lines show the mean held-out $L1$ error at each lead time and shaded areas indicate one standard deviation.}
    \label{fig:ablation_pf}
\end{figure}

Figure~\ref{fig:ablation_pf} evaluates pushforward training on top of the $Q(t)$-augmented models. On the standard mesh, E3 improves over E2 for $h$, $u$, and $v$ at all forecast horizons. The same effect appears on the multimesh graph, where E6 consistently dominates E5. This confirms that reducing the train--test mismatch during training is beneficial for long autoregressive rollouts in this boundary-driven setting. The gains become particularly visible after the first few hours, when accumulated state errors start to propagate through the graph. Taken together, Figures~\ref{fig:ablation_q}--\ref{fig:ablation_pf} suggest a cumulative progression: adding $Q(t)$ yields the largest jump in performance, multimesh connectivity brings an additional improvement once the model is properly conditioned, and pushforward further stabilises long rollouts. In state-space metrics, E6 therefore provides the strongest held-out performance among the six configurations considered in this preprint.

\subsection{Fine-grid CSI at two flood-depth thresholds}
\label{sec:results_csi_fine}
\begin{figure}[t]
    \centering
    \includegraphics[width=\linewidth]{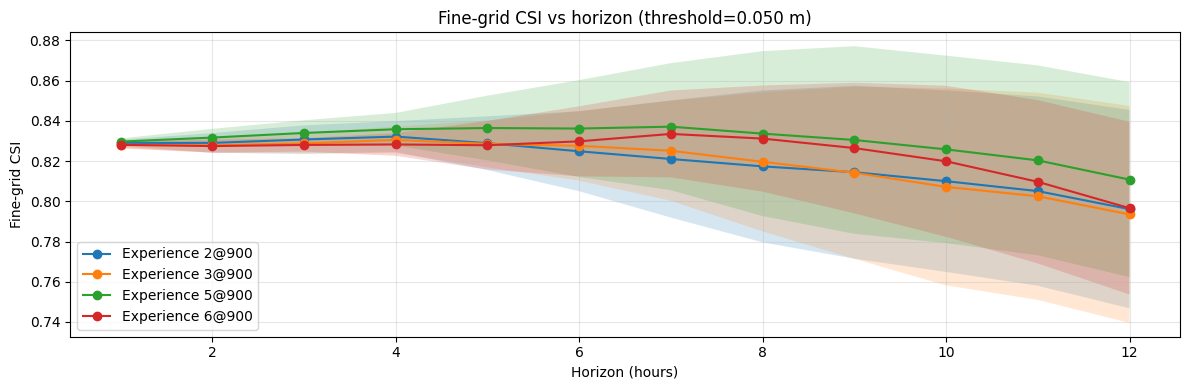}
    \caption{Fine-grid CSI on the $25\,\mathrm{m}$ common grid with threshold $h_\varepsilon=5\,\mathrm{cm}$ for the four $Q(t)$-conditioned configurations (E2, E3, E5, E6) on the 16 held-out floods. Solid lines show the mean CSI at each lead time and shaded areas indicate one standard deviation.}
    \label{fig:csi_fine_005}
\end{figure}

\begin{figure}[t]
    \centering
    \includegraphics[width=\linewidth]{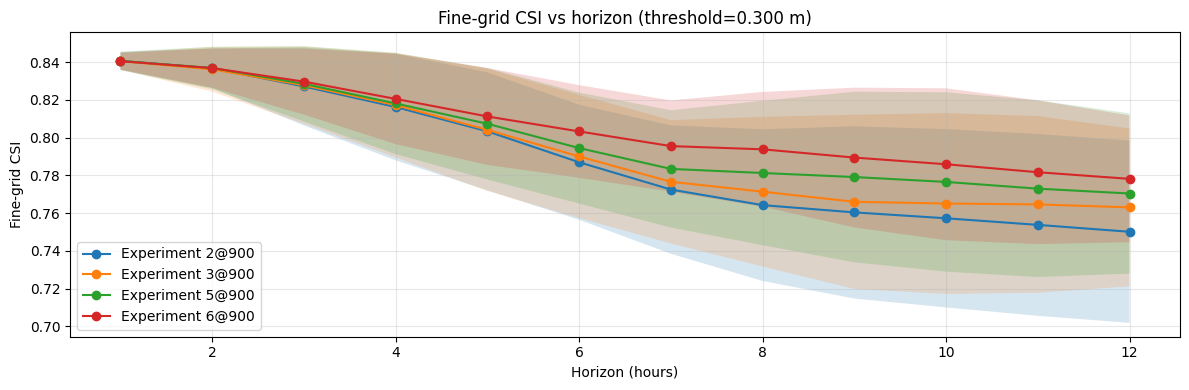}
    \caption{Fine-grid CSI on the $25\,\mathrm{m}$ common grid with threshold $h_\varepsilon=30\,\mathrm{cm}$ for the four $Q(t)$-conditioned configurations (E2, E3, E5, E6) on the 16 held-out floods. Solid lines show the mean CSI at each lead time and shaded areas indicate one standard deviation.}
    \label{fig:csi_fine_03}
\end{figure}

We now turn to map-space evaluation on the common regular grid. We use a regular support with a spacing of $25\,\mathrm{m}$. On the T\^et case study, this corresponds to a rectangular grid of $689\times748=515{,}372$ grid points, of which 239,371 are masked outside the triangulated hydraulic domain, leaving 276,001 in-domain points for the CSI computation. Importantly, the cartographic reference is obtained from the original high-resolution Telemac simulation, not from the $\times 8$ projected mesh used internally by the surrogate. Figures~\ref{fig:csi_fine_005} and \ref{fig:csi_fine_03} compare the four $Q(t)$-conditioned models using CSI computed from binary inundation maps at two depth thresholds: $h_\varepsilon=5\,\mathrm{cm}$ and $h_\varepsilon=30\,\mathrm{cm}$. These two thresholds probe different operational regimes. The $5\,\mathrm{cm}$ threshold is sensitive to the outer flood envelope and therefore to shallow marginal inundation, while the $30\,\mathrm{cm}$ threshold focuses more strongly on the hydraulically meaningful flooded core.

At $5\,\mathrm{cm}$, all four models start from a very similar CSI during the first hours, but the multimesh variants remain systematically above the standard-mesh ones as the horizon increases. The best curve is obtained by E5, with E6 remaining very close. This suggests that multimesh connectivity is particularly helpful for preserving the spatial extent of shallow inundation, while the additional pushforward training slightly trades outer-envelope coverage for rollout stability. At $30\,\mathrm{cm}$, the ranking becomes clearer: E6 dominates at essentially all horizons, E5 is second, and both standard-mesh models fall below the multimesh pair. In other words, the combination of $Q(t)$, multimesh connectivity, and pushforward is most beneficial when the target is the more decision-relevant core of the flood rather than the thinnest fringe of shallow water.

This threshold dependence is informative rather than contradictory. These results show that improvements in the prediction of the hydraulic variables also translate into better flood maps, although the benefit is not identical across depth thresholds. With a $5\,\mathrm{cm}$ threshold, which gives more weight to shallow inundation at the flood margins, E5 and E6 achieve very similar CSI values, with a slight advantage for E5. With a $30\,\mathrm{cm}$ threshold, which places more emphasis on the main flooded areas, E6 performs best at almost all lead times. We therefore regard E6 as the best overall configuration in this preprint, while noting that the most relevant evaluation threshold depends on the intended operational use of the inundation map.

\subsection{Qualitative inundation map and inference time}
\label{sec:results_qualitative}
\begin{figure}[t]
    \centering
    \includegraphics[width=\linewidth]{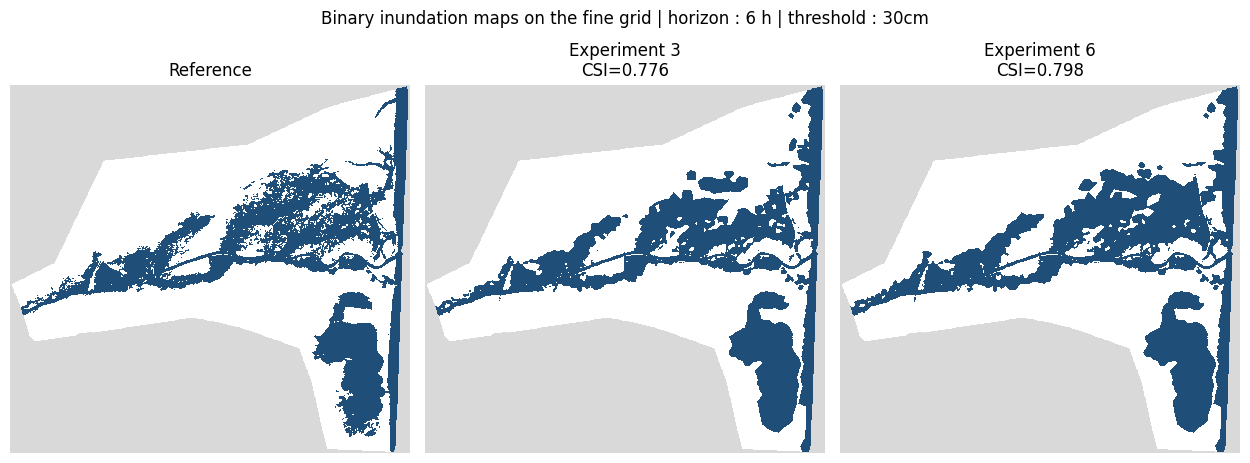}
    \caption{Qualitative binary inundation map on the $25\,\mathrm{m}$ common grid for one held-out event from hydrograph group~3 with peak discharge $3400\,\mathrm{m}^3/\mathrm{s}$, at a 6-hour horizon and threshold $h_\varepsilon=30\,\mathrm{cm}$. The reference is derived from the original high-resolution Telemac solution.}
    \label{fig:qualitative_3400}
\end{figure}

Figure~\ref{fig:qualitative_3400} shows one held-out event from hydrograph group~3 with a peak discharge of $3400\,\mathrm{m}^3/\mathrm{s}$, visualised as a binary inundation map on the common $25\,\mathrm{m}$ grid at a 6-hour horizon and with a $30\,\mathrm{cm}$ threshold. The reference map is obtained from the original high-resolution Telemac solution. Both surrogates recover the main flooded corridor and the downstream storage area, but E6 is visually closer to the reference than E3 in the central floodplain and achieves a higher CSI on this case (0.798 vs 0.776). At the same time, both learned maps remain smoother and less fragmented than the original-mesh reference, which is consistent with the aggregate CSI results discussed above.

This qualitative example also highlights the practical motivation for the surrogate approach. On our hardware, the neural rollout for this 6-hour forecast is produced in about $0.4\,\mathrm{s}$ on a single NVIDIA A100 GPU, whereas the reference prediction on the original mesh requires on average about $180\,\mathrm{min}$ on 56 CPU cores. Even allowing for the final interpolation onto the regular grid, this leaves a very large gap in walltime between the learned surrogate and the original hydrodynamic simulation.

\section{Conclusion}

This work addressed the problem of rapid flood forecasting in an operational river setting, using the lower T\^et basin as a case study. Starting from a production-grade Telemac2D model on a high-resolution unstructured finite-element mesh, we built a learning-ready database of synthetic but operationally grounded flood events, designed to cover a range of realistic hydrograph shapes and peak discharges. This database constitutes an important contribution of the present work, as it makes it possible to study graph-based surrogates in a boundary-driven setting that is much closer to operational practice than the simplified benchmarks commonly used in the literature.

On top of this database, we developed a graph-neural surrogate based on projected meshes and multimesh connectivity. The projected-mesh strategy makes learning tractable while preserving high-fidelity supervision from the original Telemac simulations, and the multimesh construction increases the effective spatial receptive field without increasing network depth. In our experiments, the discharge input $Q(t)$ proved essential to recover the timing and magnitude of flood events, while multimesh connectivity and pushforward training brought further gains in long-horizon rollout accuracy. Among the tested configurations, E6 provides the best overall compromise.

Importantly, these gains are not limited to the hydraulic variables on the surrogate mesh. They also translate into better inundation maps when predictions are compared with the original high-resolution Telemac solution on a common regular grid. This evaluation is more consistent with the intended operational product, namely flood maps that can support situational awareness and crisis management. The qualitative and quantitative results both indicate that the proposed approach can recover the main flooded structures while maintaining a very large reduction in prediction time: about $0.4\,\mathrm{s}$ on a single A100 GPU for the learned surrogate, versus about $180\,\mathrm{min}$ on 56 CPU cores for a 6-hour reference simulation.

Overall, these results support the idea that graph-based surrogates can become a practical complement to industrial hydraulic solvers for operational flood mapping. Future work will focus on extending the cartographic validation, refining the physical treatment of boundary conditions, and further strengthening the surrogate for deployment-oriented forecasting workflows. This question is particularly important in operational settings, where the inflow hydrograph is itself uncertain and may be revised over time, making robustness to boundary-forcing uncertainty a key requirement for deployment.

\section{Acknowledgement}
This work was supported by a French government grant managed by the Agence Nationale de la Recherche under the "Investissements d'avenir" program (reference "ANR-21-ESRE-0051"). The authors acknowledge NVIDIA for making the PhysicsNeMo framework publicly available. The MeshGraphNet implementation provided in this framework was used as the basis for the work presented in this article. The authors also thank Antonin Mazoyer, Fabrice Cebron, and Emmanuel Dervau from BRL Ingénierie for their valuable advice on the hydraulic aspects of this work.
\bibliographystyle{unsrt}
\bibliography{references}

\end{document}